\documentclass{article}
\usepackage{spconf,amsmath,graphicx}

\usepackage{enumitem}
\setlist{nosep, leftmargin=14pt}

\usepackage{multirow}
\usepackage{bbding}
\usepackage{pifont}
\usepackage{wasysym}
\usepackage{amssymb}
\usepackage{tabularx,ragged2e}
\newcolumntype{C}{>{\arraybackslash}X}
\usepackage{graphicx}
\usepackage{rotating}
\usepackage{graphicx}
\usepackage{booktabs}
\usepackage{multirow}
\usepackage{rotating}
\usepackage{lscape}
\usepackage{booktabs}
\usepackage{xcolor,colortbl}
\usepackage{makecell}

\usepackage{lipsum}
\usepackage{algpseudocode}
\usepackage{amsmath}
\usepackage[linesnumbered,ruled]{algorithm2e}
\definecolor{lavender}{gray}{0.9}
\usepackage{array}
\usepackage{mwe} 
\usepackage{graphicx}
\usepackage{todonotes}
\usepackage{bbding}
\usepackage{booktabs}
\usepackage{multirow,bigdelim}
\usepackage{color}
\usepackage{xr}

\usepackage{amsmath,amsfonts,amssymb}

\usepackage{rotating}
\usepackage{mwe}
\usepackage{graphbox}
\usepackage{multirow}
\usepackage{amsfonts} 
\usepackage{amsmath,mathtools}
\usepackage{amssymb}
\usepackage{bm}
\usepackage{url}
\usepackage[colorlinks=true,linkcolor=blue, citecolor=blue, urlcolor=blue]{hyperref}
\def\arrvline{\hfil\kern\arraycolsep\vline\kern-\arraycolsep\hfilneg}
\usepackage{pifont}
\newcommand\like[1]{\begin{picture}(1,1)
\ifnum0=#1\put(.5, 2.5){\circle{5}}\else
\ifnum10=#1\put(.5,2.5){\circle*{5}}\else
\put(.5,.35){\circle{0.5}}\put(.5,.35){\circle*{.0.5}}
\fi\fi\end{picture}}
\usepackage{hhline}
\usepackage{threeparttable}
\definecolor{lavender}{gray}{0.9}
\usepackage{soul}
\sethlcolor{orange!25} 
\newcommand\sbullet[1][.5]{\mathbin{\vcenter{\hbox{\scalebox{#1}{$\bullet$}}}}}

\title{Assessing Test-time Variability for Interactive 3D Medical Image Segmentation with Diverse Point Prompts}

\name{Hao Li, Han Liu, Dewei Hu, Jiacheng Wang, Ipek Oguz}
\address{Vanderbilt University} 

\begin{document}
%

\maketitle
\begin{abstract}
Interactive segmentation model leverages prompts from users to produce robust segmentation. This advancement is facilitated by prompt engineering, where interactive prompts serve as strong priors during test-time. However, this is an inherently subjective and hard-to-reproduce process. The variability in user expertise and inherently ambiguous boundaries in medical images can lead to inconsistent prompt selections, potentially affecting segmentation accuracy. This issue has not yet been extensively explored for medical imaging. In this paper, we assess the test-time variability for interactive medical image segmentation with diverse point prompts. For a given target region, the point is classified into three sub-regions: boundary, margin, and center. Our goal is to identify a straightforward and efficient approach for optimal prompt selection during test-time based on three considerations: (1) benefits of additional prompts, (2) effects of prompt placement, and (3) strategies for optimal prompt selection. We conduct extensive experiments on the public Medical Segmentation Decathlon dataset for challenging colon tumor segmentation task. We suggest an optimal strategy for prompt selection during test-time, supported by comprehensive results. The code is  publicly available at \url{https://github.com/MedICL-VU/variability} 
\end{abstract}


\section{Introduction}

To date, deep learning methods have demonstrated superior performance in various medical image segmentation tasks \cite{liu2023medical}. However, the generalizability and effectiveness of these fully automated methods may be limited by the amount of available labeled medical data \cite{wang2023ssl2}. Instead, interactive segmentation methods that integrate user knowledge and application requirements have been proposed \cite{wang2018interactive}. 


\begin{figure}[t]
\centering
\includegraphics[width=0.9\linewidth]{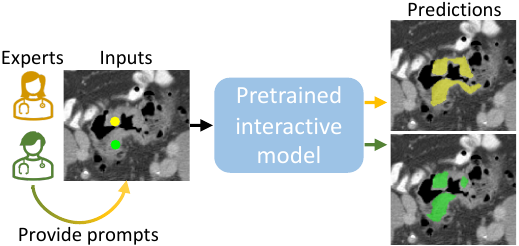}   
\caption{A 3D interactive segmentation model, which takes an image and a point prompt as inputs. The segmentation result can vary with different prompts provided by experts. 
}
\label{variability}
\end{figure}

\begin{figure*}[t]
\centering
\includegraphics[width=0.93\linewidth]{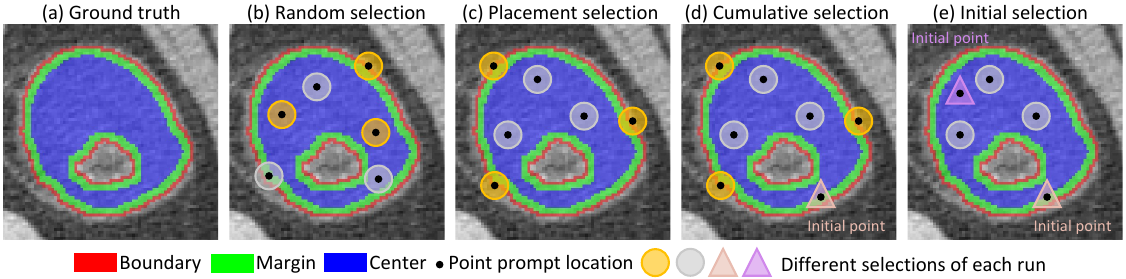}   
\caption{Various point prompt selection strategies. Different prompt colors represent different runs. \textbf{(a)} Ground truth, highlighting the entire tumor and its three different sub-regions. \textbf{(b)} Random selection throughout the entire tumor. \textbf{(c)} Prompts confined to a specific sub-region. \textbf{(d)} Cumulative selection, where the initial point (\underline{triangle}) is fixed while cumulative points (\underline{circles}) vary between runs. \textbf{(e)} Initial selection, where initial points vary while cumulative points remain fixed. }
\label{point_sampling}
\end{figure*}

To tackle the data and label availability issue in medical imaging, it is desirable to utilize knowledge from the natural images domain, wherein large public datasets are more readily accessible \cite{matsoukas2022makes}. Recent studies \cite{ma2023segment,wu2023medical,gong20233dsam,li2023promise,deng2023sam,zhang2023input,zhang2023segment} have leveraged insights from pretrained natural image foundation models, such as the Segment Anything Model (SAM) \cite{kirillov2023segment}, which is trained on massive datasets, to facilitate robust medical image segmentation through parameter-efficient transfer learning techniques. These methods, which use diverse visual prompts, achieve robust performance due to the strong prior provided by the interactive prompt during test-time.

Compared to other interaction formats, such as boxes or scribbles, point prompts are preferable in practice, as they require less effort, especially for 3D medical images. However, in the context of interactive segmentation models, determining precise key points during inference can be elusive, especially in medical images characterized by low quality, poor contrast, and ambiguous boundaries. Furthermore, subjectivity leads to variability in prompt choice as seen in Fig.~\ref{variability}, and this can be exacerbated by different user expertise levels, resulting in different segmentation outcomes at test-time \cite{schmidt2023probabilistic,yao2023false,cheng2023sam}. A random selection strategy is widely used, but random points might not represent the key features of a medical image effectively. To the best of our knowledge, no existing work has explicitly addressed prompt selection in the medical field. Previous works have highlighted the importance of key points and have leveraged these during training to produce robust segmentations \cite{lin2020interactive,lee2020structure}. 
In this paper, we aim to investigate the optimal strategy for point prompt selection for pretrained interactive segmentation models at test-time.

We first assess the test-time variability caused by point prompt selection of interactive 3D medical image segmentation model. 
Specifically, we use the ProMISe \cite{li2023promise} model as the backbone, which leverages the pretrained weights from SAM \cite{kirillov2023segment}. It 
takes image and point prompts as inputs to produce a segmentation. We evaluate the segmentation performance variability caused by different number, region, and selection strategy for prompts during inference to quantify how such interactive models respond to diverse point prompts (Fig.~\ref{point_sampling}). To provide a pragmatic solution, our investigation focuses on three aspects: (1) determining the necessary number of prompts, (2) identifying effective prompt placement locations, and (3) formulating a strategy for prompt selection.

We conduct our evaluation on colon tumor segmentation from the public Medical Segmentation Decathlon dataset \cite{antonelli2022medical}, which is characterized by irregular shapes and ambiguous boundaries. Our findings suggest a straightforward strategy for test-time prompt selection without requiring additional effort, leading to significantly improved segmentation results over random prompt selection.



\section{Methods}

\noindent\textbf{Dataset.} We used the Medical Segmentation Decathlon \cite{antonelli2022medical} for our experiments. 
We select the challenging colon tumor segmentation task where ambiguous edges and irregular shape are present. The dataset contains 126 3D CT volumes with resolution ranging from $0.54\times0.54\times1.25$ to $0.98\times0.98\times7.5 mm^3$, resampled to 1$mm$ isotropic resolution.
The training, validation and test sets contain 88, 12 and 26 subjects, respectively. 
 We perform intensity clipping based on foreground $0.5$ and $99.5$ percentiles, and Z-score normalization based on foreground voxels of training set. 


\noindent\textbf{Interactive segmentation model.} We employ ProMISe \cite{li2023promise}, a SAM-based interactive segmentation model which takes an input image and point prompts as inputs. During training, an input patch of size $128\times128\times128$ was randomly selected such that its center voxel is equally likely to be foreground or background. In addition, random flip, rotation, zoom and intensity shift are used as data augmentation strategies. For point prompts, 10 positive points from the foreground and 20 negative points from the background are randomly selected for each input patch, respectively. We use negative points only if the number of positive points in an input patch is fewer than 10. During inference, ProMISe takes the input image and only a single random point prompt within the whole tumor area to produce the segmentation (Fig.~\ref{variability}). We consider this random prompt selection as the baseline method and evaluate various other point prompt selection strategies, aiming to identify a better strategy to optimize the segmentation performance without requiring extensive additional user effort.

\begin{table*}[t]
\centering
\sethlcolor{blue!20} 
\caption{Quantitative results of random selection, presented as ${mean}\pm{std.\ dev}$.\kern 0.7em \like{0} \space and \space\like{10} \space denote absent and present for the point prompt select region.\kern 0.3em \textbf{Bold} and \underline{underline} denote the best  performances for each \textbf{column (region-wise)} and \underline{row (prompt-wise)}, respectively. \hl{Blue} indicates the baseline method. \\}
\small	
\label{main_table}
\setlength{\tabcolsep}{3.5pt}
\begin{tabular}{ccc | ccccc|ccccc}
\hline
\multicolumn{3}{c}{Region} & \multicolumn{5}{c}{Dice score} & \multicolumn{5}{c}{Normalized surface Dice} \\ \hline
\multicolumn{1}{c}{$B$} & \multicolumn{1}{c}{$M$} & \multicolumn{1}{c|}{$C$} 

& \multicolumn{1}{c}{$1P$} & \multicolumn{1}{c}{$5P$} & \multicolumn{1}{c}{$10P$} & \multicolumn{1}{c}{$20P$} & \multicolumn{1}{c|}{$100P$}

& \multicolumn{1}{c}{$1P$} & \multicolumn{1}{c}{$5P$} & \multicolumn{1}{c}{$10P$} & \multicolumn{1}{c}{$20P$} & \multicolumn{1}{c}{$100P$}  \\
\hline

\rowcolor{lavender}
\like{10} & \like{0} & \like{0} 
& .622$\pm$.014  & .650$\pm$.009 &  .652$\pm$.008 & .650$\pm$.007 & \textbf{\underline{.655$\pm$.006}} 

& .768$\pm$.015 & .798$\pm$.010 &  .803$\pm$.009 & .801$\pm$.009 & \underline{.806$\pm$.007} \\

\like{0} & \like{10} & \like{0} 
& .630$\pm$.015 & .652$\pm$.010 &  \underline{.654$\pm$.009} &  .652$\pm$.008  & \underline{.654$\pm$.006}  

& .776$\pm$.016 & .802$\pm$.011 &  .805$\pm$.009  & .805$\pm$.009  & \textbf{\underline{.807$\pm$.008}} \\

\rowcolor{lavender}
\like{0} & \like{0} & \like{10} 
& \textbf{.642$\pm$.012} & \textbf{\underline{.654$\pm$.006}}  &  \underline{.654$\pm$.006}  & .652$\pm$.009  & .653$\pm$.005  

& \textbf{.788$\pm$.015} & .802$\pm$.006  &  .804$\pm$.007  & .803$\pm$.010 &  \underline{.805$\pm$.007} \\

\like{10} & \like{10} & \like{0} 
& .632$\pm$.014 & .653$\pm$.010 & .652$\pm$.008 & \textbf{\underline{.654$\pm$.007}} & .652$\pm$.005

& .777$\pm$.016 & .803$\pm$.011 & .802$\pm$.010 & \textbf{\underline{.806$\pm$.007}} & .804$\pm$.007\\

\rowcolor{lavender}
\like{10} & \like{0} & \like{10} 
& .634$\pm$.016 & .652$\pm$.010 & .653$\pm$.009 & \underline{.653$\pm$.008} & .652$\pm$.005

& .779$\pm$.017 & .801$\pm$.010 & .803$\pm$.010 & .804$\pm$.009 & \underline{.805$\pm$.007} \\

\like{0} & \like{10} & \like{10} 
& .634$\pm$.016 & .654$\pm$.009 & .654$\pm$.008 & .653$\pm$.009 & \underline{.654$\pm$.007}

& .779$\pm$.017 & .803$\pm$.011 & .804$\pm$.009 & .805$\pm$.010 & \underline{.807$\pm$.009} \\

\rowcolor{lavender}
\like{10} & \like{10} & \like{10} 
& \cellcolor{blue!20}.637$\pm$.014  & .653$\pm$.008  & \textbf{\underline{.655$\pm$.008}}   & .653$\pm$.007  & .652$\pm$.007  

& .783$\pm$.016  & \textbf{.803$\pm$.008}  & \textbf{\underline{.806$\pm$.009}}   & \textbf{\underline{.806$\pm$.007}}  & .804$\pm$.008 \\

\hline
\multicolumn{13}{l}{Abbreviations. $B$=boundary, $M$=margin, $C$=center, $P$=point prompt(s) per volume.}
\end{tabular}
\end{table*}
\begin{table*}[t]
\caption{Quantitative results of cumulative selection.\kern 0.3em \textbf{Bold} and \underline{underline} denote the best  performances for each \textbf{column (region-wise)} and \underline{row (prompt-wise)}, respectively. \hl{Orange} indicates the suggested selection strategy.}
\begin{center}
\label{cumulative}
\setlength{\tabcolsep}{3.5pt}
\begin{tabular}{c|ccc | cccc|cccc}
\hline
\multicolumn{4}{c}{Region} & \multicolumn{8}{c}{(Initial $+$ cumulative) points} \\ 
\cmidrule(lr){1-4} \cmidrule(lr){5-12}

\multicolumn{1}{c}{$Init.$} & \multicolumn{3}{c}{$Cumu.$} & \multicolumn{4}{c}{Dice score} & \multicolumn{4}{c}{Normalized surface Dice} \\ \hline

\multicolumn{1}{c|}{$W$} & \multicolumn{1}{c}{$B$} & \multicolumn{1}{c}{$M$} & \multicolumn{1}{c|}{$C$} 

& \multicolumn{1}{c}{$(1+4)P$} & \multicolumn{1}{c}{$(5+5)P$} & \multicolumn{1}{c}{$(10+10)P$} & \multicolumn{1}{c|}{$(20+80)P$}

& \multicolumn{1}{c}{$(1+4)P$} & \multicolumn{1}{c}{$(5+5)P$} & \multicolumn{1}{c}{$(10+10)P$} & \multicolumn{1}{c}{$(20+80)P$}  \\
\hline

\like{10} & \like{10} & \like{0} & \like{0} 
& .651$\pm$.011  & .652$\pm$.008 &  .652$\pm$.007 & \textbf{\underline{.655$\pm$.005}}  

& .799$\pm$.009 & .803$\pm$.009 &  .804$\pm$.008 & \textbf{\underline{.807$\pm$.007}} \\

\rowcolor{lavender}
\like{10} & \like{0} & \like{10} & \like{0} 
& .654$\pm$.011  & \underline{.654$\pm$.008} &  .653$\pm$.008 & .653$\pm$.005  

& .804$\pm$.009 & \textbf{\underline{.806$\pm$.009}} &  \underline{.806$\pm$.009} & .806$\pm$.007 \\

\like{10} & \like{0} & \like{0} & \like{10} 
& \cellcolor{orange!25}\textbf{\underline{.657$\pm$.008}}  & \textbf{.654$\pm$.007} &  \textbf{.655$\pm$.007} & .653$\pm$.007  

& \textbf{.805$\pm$.008} & .804$\pm$.008 &  \textbf{\underline{.808$\pm$.009}} & .806$\pm$.008 \\

\rowcolor{lavender}
\like{10} & \like{10} & \like{10} & \like{10} 
& .654$\pm$.010  & \textbf{\underline{.654$\pm$.007}} &  .653$\pm$.007 & .652$\pm$.007  

& .804$\pm$.010 & \underline{.805$\pm$.008} &  \underline{.805$\pm$.008} & .804$\pm$.008 \\

\hline
\multicolumn{12}{l}{\small Abbreviations. $W$=whole, $B$=boundary, $M$=margin, $C$=center, $P$=point prompt(s) per volume, $Init.$=Initial, $Cumu.$=cumulative.}
\vspace{-1em}
\end{tabular}
\end{center}
\end{table*}

\noindent\textbf{Sub-region generation.} As shown in Fig.~\ref{point_sampling}(a),  we divide the ground truth mask into three distinct sub-regions: boundary, margin, and center. We generate these regions using an average-pooling operation, denoted as $P_{ave}^k$, where $k$ represents the kernel size. For a given a binary segmentation $S$, the binary boundary sub-region is derived: $B(S) = thres ( S \odot |S - P_{ave}^3(S)| )$, where $|\sbullet|$ denote the absolute value function, $\odot$ represents element-wise multiplication, and $thres$ is binary thresholding with a threshold of 0. Similarly, we obtain the margin sub-region as: $M(S) = thres ( S \odot |S - P_{ave}^7(S)| ) - B(S)$, and the center sub-region is simply $C(S) = S - M(S) - B(S)$. This sub-region generation can be seamlessly integrated into the inference without reducing computation speed, and may be used to facilitate pseudo-label learning.


\noindent\textbf{Random selection.} To ensure objectivity and fair comparison, random selection serves as the underlying method in our approach for every selection strategy, instead of actual user prompts. Additionally, we use randomly selecting single point prompts within the whole tumor area as the baseline method for comparison, as this represents the most common selection strategy. We specify the random seed to control variability and assess the impact of different settings within these selection strategies. It is important to note that the location of point prompts depends on the seed and the selected region. In other words, for a specific region and seed, the point locations remain the same if the number of point prompts is unchanged, and differences only arise with additional points. Moreover, we investigate the benefits of using more than one prompt. Specifically, we use 1, 5, 10, 20, and 100 prompts. 

\noindent\textbf{Placement selection.} 
To identify the effectiveness of point placement, the point prompts are randomly selected within the given regional constraints. Fig.~\ref{point_sampling}(c) depicts a situation with a same seed setting but different selection region.


\noindent\textbf{Strategy selection.} The initial selection is important in prompt selection \cite{lin2020interactive}. To further examine its test-time variability, we assess two strategy selections, cumulative and initial selections, as shown in Figs.~\ref{point_sampling}(d) and (e), respectively. For cumulative selection, we randomly set an initial point (triangle in Fig.~\ref{point_sampling}(d)) in a specific region as the fixed point, and then compare the segmentation performance for different placements of the cumulative points (circles). 
All selected prompts, including initial and cumulative points, are used for each run. Similarly, in Fig.~\ref{point_sampling}(e), we select different initial points while keeping the cumulative points fixed. These approaches are practical since selecting a single point prompt is preferred in practice for its minimal effort, compared to other prompt types such as boxes, scribbles or multiple point prompts. Furthermore, a preliminary segmentation or pseudo label may first be generated from the initial point(s) to serve as a reference for the selection of the cumulative points.

\renewcommand{\thefigure}{4}
\begin{figure*}[t]
\centering
\includegraphics[width=0.9\linewidth]{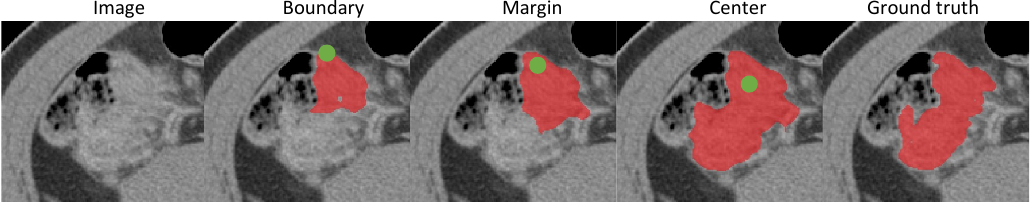}   
\caption{Qualitative results for random selection with a single point prompt (green). Labels show the selection regions.}
\label{qualitative}
\end{figure*}

\noindent\textbf{Implementation details.}
We trained the ProMISe \cite{li2023promise} model 
for a maximum of 200 epochs with batch size of 1. The initial learning rate was 0.0004, decreasing by $2e^{-6}$ every epoch, and the AdamW optimizer was employed.
During inference, we randomly selected a set of 50 random seeds and used it for each selection strategy. The Dice score and normalized surface Dice were used for evaluation, and we report  mean and standard deviations over the 50 runs (seeds). 
We used PyTorch, MONAI and an NVIDIA A6000 for our experiments.

\section{Results}




\noindent\textbf{Quantitative results.} Tab.~\ref{main_table} presents a detailed comparison of random selections, focusing mainly on two practical questions: whether more points are needed, and where to select them. The results clearly show that using more than a single prompt benefits segmentation in both metrics compared to  a single point prompt. However, a distinct cutoff in improvement is observed at five point prompts per 3D volume, which suggests diminishing returns past that. Compared to the baseline method, choosing a single point prompt focusing on the center region often yields superior segmentation. As the number of points increases, the impact of choosing different regions becomes less pronounced.

\begin{table}[h]

\caption{Quantitative results of initial selection. The \textbf{bold} and \underline{underline} denoted best performances for each \textbf{column (region-wise)} and \underline{row (prompt-wise)}, respectively. \hl{Orange} indicates the suggested selection strategy.}
\begin{center}
\small	
\setlength{\tabcolsep}{3.5pt}
\label{initial}
\begin{tabular}{ l c c c c c}
\hline
\multicolumn{2}{c}{} & \multicolumn{4}{c}{Region (Dice)} \\ 
 \cmidrule(lr){3-6}





\multicolumn{1}{c}{$Init.$} & $Cumu.$ & $B$ & $M$ & $C$ & $W$ \\ \hline
1$(W)$  & 4$P$ &  .651$\pm$.011 &  .654$\pm$.011 & \cellcolor{orange!25}\textbf{\underline{.657$\pm$.008}}   &  .654$\pm$.010 \\

\rowcolor{lavender}
1$(W)$ & 9$P$ &  .652$\pm$.008 &  .655$\pm$.008 & \textbf{\underline{.655$\pm$.006}}   &  .653$\pm$.010 \\

\hline

1$(C)$ & 4$P$ &  .653$\pm$.008 & .651$\pm$.007 & \textbf{\underline{.654$\pm$.006}}  &  .652$\pm$.008\\

\rowcolor{lavender}
1$(C)$ & 9$P$ &  .654$\pm$.007 & .654$\pm$.007 & \textbf{\underline{.654$\pm$.006}}  &  .653$\pm$.009\\

\hline

\end{tabular}
\end{center}
\end{table}

\renewcommand{\thefigure}{3}
\begin{figure}[h]
\centering
\includegraphics[width=0.93\linewidth]{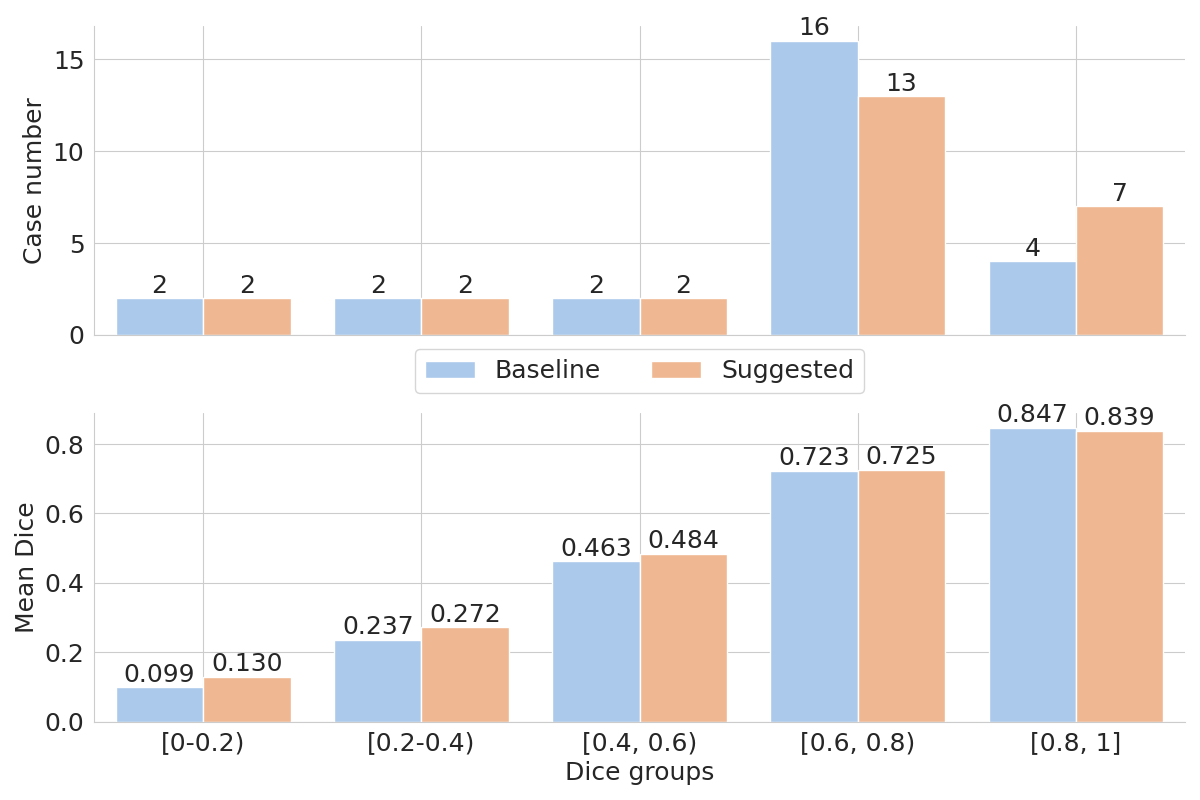}   
\caption{Comparison of performance on Dice distribution.}
\label{dice_group}
\end{figure}

Tab.~\ref{cumulative} compares the different cumulative strategies with different cumulative point placement and different number of prompts. We found that randomly selecting a single point prompt in the whole tumor area, with cumulative points  picked from the center region achieves the best results. This improves the Dice score of baseline method by 2\%, with statistical significance confirmed ($p<0.001$) through a 2-tailed paired t-test. Therefore, we suggest this straightforward selection strategy as the optimal solution during test-time. We note that the cumulative points from the easily identifiable center area can either be derived from a pseudo label or require minimal extra effort by experts to improve segmentation performance. Thus, in practice, selecting a single initial point is enough for the recommended method, resulting in good performance for minimal prompt input effort.

\renewcommand{\thefigure}{5}
\begin{figure}[h]
\centering
\includegraphics[width=0.87\linewidth]{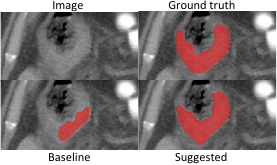}   
\caption{Qualitative results of suggested and baseline methods.}
\label{suggestion}
\end{figure}

Tab.~\ref{initial} shows the impact of initial selection region. The results indicate that the whole and center areas are the best regions for initial and cumulative points, respectively. However, the differences among the methods are minor.


Fig.~\ref{dice_group} presents the impact of suggested method on different subjects grouped by Dice. Compared to baseline method, the suggested method has higher Dice scores for most groups. In addition, the suggested method has more contribution to the subjects with high-quality segmentations produced by baseline method, as evidenced by Dice scores and case numbers for the two highest Dice groups.

\noindent\textbf{Qualitative results.} 
Fig.~\ref{qualitative} shows qualitative results for the random selection with a single point prompt. Both boundary and margin selections produce under-segmented results near the ambiguous areas, while center selection captures the missing areas. In Fig.~\ref{suggestion}, the suggested method dramatically improves segmentation. The results match the ground truth well, with the exception of slightly over-segmented areas.

\section{conclusion}

In this paper, we suggest a straightforward prompt  strategy for interactive test-time segmentation, without requiring extensive additional effort. We evaluate  on a public dataset for the challenging colon tumor segmentation task, with a significant improvement over the baseline method. Future work will validate the suggested method on more public datasets. 

\noindent\textbf{Acknowledgments.}
This work was supported, in part, by NIH U01-NS106845, and NSF grant 2220401.

\clearpage


\clearpage
\section{Compliance with Ethical Standards}
This research study was conducted retrospectively using human subject data made available in open access by MSD. Ethical approval was not required as confirmed by the license attached with the open access data.

\bibliographystyle{IEEEbib}
\bibliography{refs}

\begin{thebibliography}{10}

\bibitem{liu2023medical}
Han Liu, Dewei Hu, Hao Li, and Ipek Oguz,
\newblock ``Medical image segmentation using deep learning,''
\newblock in {\em Machine Learning for Brain Disorders}, pp. 391--434. Springer, 2023.

\bibitem{wang2023ssl2}
Jiacheng Wang, Hao Li, Han Liu, Dewei Hu, Daiwei Lu, Keejin Yoon, Kelsey Barter, Francesca Bagnato, and Ipek Oguz,
\newblock ``Ssl2 self-supervised learning meets semi-supervised learning: multiple clerosis segmentation in 7t-mri from large-scale 3t-mri,''
\newblock in {\em Medical Imaging 2023: Image Processing}. SPIE, 2023, vol. 12464, pp. 126--136.

\bibitem{wang2018interactive}
Guotai Wang, Wenqi Li, Maria~A Zuluaga, Rosalind Pratt, Premal~A Patel, Michael Aertsen, Tom Doel, Anna~L David, Jan Deprest, S{\'e}bastien Ourselin, et~al.,
\newblock ``Interactive medical image segmentation using deep learning with image-specific fine tuning,''
\newblock {\em IEEE transactions on medical imaging}, vol. 37, no. 7, pp. 1562--1573, 2018.

\bibitem{matsoukas2022makes}
Christos Matsoukas, Johan~Fredin Haslum, Moein Sorkhei, Magnus S{\"o}derberg, and Kevin Smith,
\newblock ``What makes transfer learning work for medical images: Feature reuse \& other factors,''
\newblock in {\em Proceedings of the IEEE/CVF Conference on Computer Vision and Pattern Recognition}, 2022, pp. 9225--9234.

\bibitem{ma2023segment}
Jun Ma and Bo~Wang,
\newblock ``Segment anything in medical images,''
\newblock {\em arXiv preprint arXiv:2304.12306}, 2023.

\bibitem{wu2023medical}
Junde Wu, Rao Fu, Huihui Fang, Yuanpei Liu, Zhaowei Wang, Yanwu Xu, Yueming Jin, and Tal Arbel,
\newblock ``Medical sam adapter: Adapting segment anything model for medical image segmentation,''
\newblock {\em arXiv preprint arXiv:2304.12620}, 2023.

\bibitem{gong20233dsam}
Shizhan Gong, Yuan Zhong, Wenao Ma, Jinpeng Li, Zhao Wang, Jingyang Zhang, Pheng-Ann Heng, and Qi~Dou,
\newblock ``3dsam-adapter: Holistic adaptation of sam from 2d to 3d for promptable medical image segmentation,''
\newblock {\em arXiv preprint arXiv:2306.13465}, 2023.

\bibitem{li2023promise}
Hao Li, Han Liu, Dewei Hu, Jiacheng Wang, and Ipek Oguz,
\newblock ``Promise: Prompt-driven 3d medical image segmentation using pretrained image foundation models,''
\newblock {\em arXiv preprint arXiv:2310.19721}, 2023.

\bibitem{deng2023sam}
Guoyao Deng, Ke~Zou, Kai Ren, Meng Wang, Xuedong Yuan, Sancong Ying, and Huazhu Fu,
\newblock ``Sam-u: Multi-box prompts triggered uncertainty estimation for reliable sam in medical image,''
\newblock {\em arXiv preprint arXiv:2307.04973}, 2023.

\bibitem{zhang2023input}
Yizhe Zhang, Tao Zhou, Peixian Liang, and Danny~Z Chen,
\newblock ``Input augmentation with sam: Boosting medical image segmentation with segmentation foundation model,''
\newblock {\em arXiv preprint arXiv:2304.11332}, 2023.

\bibitem{zhang2023segment}
Yichi Zhang and Rushi Jiao,
\newblock ``Towards segment anything model (sam) for medical image segmentation: A survey,'' 2023.

\bibitem{kirillov2023segment}
Alexander Kirillov, Eric Mintun, Nikhila Ravi, Hanzi Mao, Chloe Rolland, Laura Gustafson, Tete Xiao, Spencer Whitehead, Alexander~C Berg, Wan-Yen Lo, et~al.,
\newblock ``Segment anything,''
\newblock {\em arXiv preprint arXiv:2304.02643}, 2023.

\bibitem{schmidt2023probabilistic}
Arne Schmidt, Pablo Morales-{\'A}lvarez, and Rafael Molina,
\newblock ``Probabilistic modeling of inter-and intra-observer variability in medical image segmentation,''
\newblock in {\em Proceedings of the IEEE/CVF International Conference on Computer Vision}, 2023, pp. 21097--21106.

\bibitem{yao2023false}
Xing Yao, Han Liu, Dewei Hu, Daiwei Lu, Ange Lou, Hao Li, Ruining Deng, Gabriel Arenas, Baris Oguz, Nadav Schwartz, et~al.,
\newblock ``False negative/positive control for sam on noisy medical images,''
\newblock {\em arXiv preprint arXiv:2308.10382}, 2023.

\bibitem{cheng2023sam}
Dongjie Cheng, Ziyuan Qin, Zekun Jiang, Shaoting Zhang, Qicheng Lao, and Kang Li,
\newblock ``Sam on medical images: A comprehensive study on three prompt modes,''
\newblock {\em arXiv preprint arXiv:2305.00035}, 2023.

\bibitem{lin2020interactive}
Zheng Lin, Zhao Zhang, Lin-Zhuo Chen, Ming-Ming Cheng, and Shao-Ping Lu,
\newblock ``Interactive image segmentation with first click attention,''
\newblock in {\em Proceedings of the IEEE/CVF conference on computer vision and pattern recognition}, 2020, pp. 13339--13348.

\bibitem{lee2020structure}
Hong~Joo Lee, Jung~Uk Kim, Sangmin Lee, Hak~Gu Kim, and Yong~Man Ro,
\newblock ``Structure boundary preserving segmentation for medical image with ambiguous boundary,''
\newblock in {\em Proceedings of the IEEE/CVF conference on computer vision and pattern recognition}, 2020, pp. 4817--4826.

\bibitem{antonelli2022medical}
Michela Antonelli, Annika Reinke, Spyridon Bakas, Keyvan Farahani, Annette Kopp-Schneider, Bennett~A Landman, Geert Litjens, Bjoern Menze, Olaf Ronneberger, Ronald~M Summers, et~al.,
\newblock ``The medical segmentation decathlon,''
\newblock {\em Nature communications}, vol. 13, no. 1, pp. 4128, 2022.

\end{thebibliography}

\end{document}